\documentclass[11pt]{article}
\usepackage{coling2018}
\usepackage{times}
\usepackage{url}
\usepackage{latexsym}

\usepackage[utf8]{inputenc}
\usepackage{CJKutf8}
\usepackage{tabularx}
\usepackage{array}
\usepackage{amssymb}
\usepackage{amsmath}
\usepackage{graphicx}

\title{Enhancing Chinese Intent Classification by Dynamically Integrating Character Features into Word Embeddings with Ensemble Techniques}

\begin{document}
\author{Ruixi Lin, Charles Costello, Charles Jankowski\\CloudMinds Technology Inc\\Santa Clara, CA, USA\\ \{ruixi.lin, charlie.costello, charles.jankowski\}@cloudminds.com}
\maketitle
\begin{abstract}
  Intent classification has been widely researched on English data with deep learning approaches that are based on neural networks and word embeddings. The challenge for Chinese intent classification stems from the fact that, unlike English where most words are made up of 26 phonologic alphabet letters, Chinese is logographic, where a Chinese character is a more basic semantic unit that can be informative and its meaning does not vary too much in contexts. Chinese word embeddings alone can be inadequate for representing words, and pre-trained embeddings can suffer from not aligning well with the task at hand. To account for the inadequacy and leverage Chinese character information, we propose a low-effort and generic way to dynamically integrate character embedding based feature maps with word embedding based inputs, whose resulting word-character embeddings are stacked with a contextual information extraction module to further incorporate context information for predictions. On top of the proposed model, we employ an ensemble method to combine single models and obtain the final result. The approach is data-independent without relying on external sources like pre-trained word embeddings. The proposed model outperforms baseline models and existing methods.
\end{abstract}

\blfootnote{
    %
    % for review submission
    %
    \hspace{-0.65cm}  % space normally used by the marker
    Place licence statement here for the camera-ready version.
    % 
    % % final paper: en-us version 
    %
    % \hspace{-0.65cm}  % space normally used by the marker
    % This work is licensed under a Creative Commons 
    % Attribution 4.0 International License.
    % License details:
    % \url{http://creativecommons.org/licenses/by/4.0/}
}

\section{Introduction}
The task of multiclass user intent classification comes from the background of conversational agents, like chatbots. For example, when a chatbot system processes a user query, the first step is to identify the user intent.

The challenge for Chinese intent classification stems from the fact that, unlike English where most words are made up of 26 phonologic alphabet letters, Chinese is logographic. Chinese words are composed of Chinese characters which are logograms that have independent meanings and the meanings vary in contexts. Previous works on Chinese intent classification mainly adopt pre-trained word embedding vectors for learning, however, compared to other text classification tasks, in intent detection, the text contains more low-frequency domain-specific words like flight number or the name of a dish, which are out of vocabulary for many pre-trained word embeddings. Those less frequent words can share common characters with the more frequent words, like \begin{CJK*}{UTF8}{gbsn}``步行街''\end{CJK*}  and \begin{CJK*}{UTF8}{gbsn}``步行''\end{CJK*}  (“walkway” and “walk”), but with embedding learning tools like Word2Vec, the commonality of morphology between \begin{CJK*}{UTF8}{gbsn}``步行街''\end{CJK*} and \begin{CJK*}{UTF8}{gbsn}``步行''\end{CJK*} are lost since they are converted to different word ids. On the other hand, Chinese characters occur more frequently in fixed collocations, which limits the different contexts around a character, and this would make training Chinese character embeddings easier and more accurate, and hence features learned from Chinese characters are very informative. For an analogy, the close counterpart to Chinese characters is English subwords, like suffixes and prefixes, and Chinese radicals (graphical components of a Chinese character) are the close counterpart to English characters. Therefore, incorporating character or character n-gram vectors into the conventional word vectors can help capture the morphology within a rare Chinese word to produce a better vector representation, because the character embeddings can be shared across rare and frequent words.

In addition, the meaning and relevancy of a word to the conversational intent are closely related to the sentence context it is in, but fixed pre-trained word and character embedding vectors are unable to adapt to contextual information. To account for the lack of adaptivity, word and character embeddings could be dynamically updated during training.

To address the inadequacy of using static, pre-trained word embeddings alone for intent classification, we propose a CNN based approach to learn dynamic character-level n-gram feature maps, which is integrated with word embedding vectors during classifier training and real-time inference, the word-character embeddings obtained are stacked with an LSTM based contextual information extraction module to produce a final class probability distribution. The highlights of the experimental results are summarized as follows:

\begin{itemize}
\item A 2-D CNN based approach to learn dynamic character-level n-gram feature maps and integrate dynamically with word embedding vectors. We have observed that the proposed models outperformed models without utilizing characters by run experiments on a Chinese benchmark dataset.
\item Combining multiple proposed word-character models with ensemble techniques, we have observed an F1 score of 93.55\% on the Chinese SMP benchmark dataset, which outperforms some existing methods and is on par with the state-of-the-art result of 93.91\%, while greatly saving development time.
\item The word-character approach can be applied with low efforts to many kinds of existing neural network models simply by inserting the proposed word-character embedding module.
\end{itemize}

The remainder of this paper is organized in the following way. In related work section we present works related to intent classification and embedding learning. In model section, we detail the word-character embedding integration approach and the overall architecture, with ensemble methods used to combine our single models. This is followed by experimental setup, and results and discussion sections, where we describe experimental settings, then show, compare, and analyze experimental results. In the last section, we conclude on the works we have done and discuss future works.

\section{Related Work}
We review works related to the proposed method, including intent classification methods and works on joint word and character embeddings.

Intent classification has been an ongoing topic of research in spoken language understanding
\cite{Mori2007,Bechet2008,Tur2010}. Previous works have been done on machine learning methods like Support Vector Machines (SVMs) and Conditional Random Fields (CRFs), n-gram language models, or combined approaches for intent determination \cite{Wang2002,Wu2005,Raymond2007}. Knowledge based methods have also been explored \cite{Broder2007,Li2008,Hu2009}.

In recent years, neural network based architectures and word embeddings have gained growing popularity for intent classification. Recurrent Neural Network Language Models (RNN LMs) have been proved to be effective for capturing temporal sentence semantics through a series of hidden units (\cite{Mikolov2010}). Long Short-Term Memory (LSTM) neural networks were developed to further avoid the exploding and vanishing gradient problem of traditional RNNs by regulating the memory cells through activation computed by the gates \cite{Hochreiter1997,Gers2000}. Deep RNNs and LSTMs based models with word embeddings have shown remarkable results on slot and intent detection, and there are models that jointly detects both \cite{Yao2013,Yao2014,Ravuri2015,Shi2015,Shi2016,Zhang2016}. Convolutional Neural Network (CNN) based CRFs have also been applied and shown comparable results on joint detection of slots and intents \cite{Xu2013}.

Moving from using word embeddings alone to richer semantic representations, utilizing  character embeddings to represent words \cite{Kim2016,Bojanowski2017}, leveraging external lexicons like WordNet to enrich word embeddings \cite{Kim20161} are some of the effective approaches to replace the previous word-level input. Another approach is to incorporate character-level embeddings into word embeddings, which has been shown useful for Part-of-Speech tagging, named entity recognition, and  implicit discourse relation recognition \cite{Santos2014,Santos2015,Qin2016}. Joint learning of word and character embeddings have also drawn growing attention, where the character-enhanced word embeddings are evaluated against intrinsic word-level tasks including word relatedness and word analogy, and can be further used for extrinsic NLP tasks such as text classification \cite{Xu2016,Chen2015,Yu2017}. For Chinese related work particularly, sub-character information has also been leveraged. Radical-level character learning methods proposed by \cite{Sun2014,Li2015} are proved to work well on either Chinese character similarity judgment and Chinese word segmentation, or word similarity and text classification.

\section{Model}
\begin{figure}[t]

  \includegraphics[width=16cm]{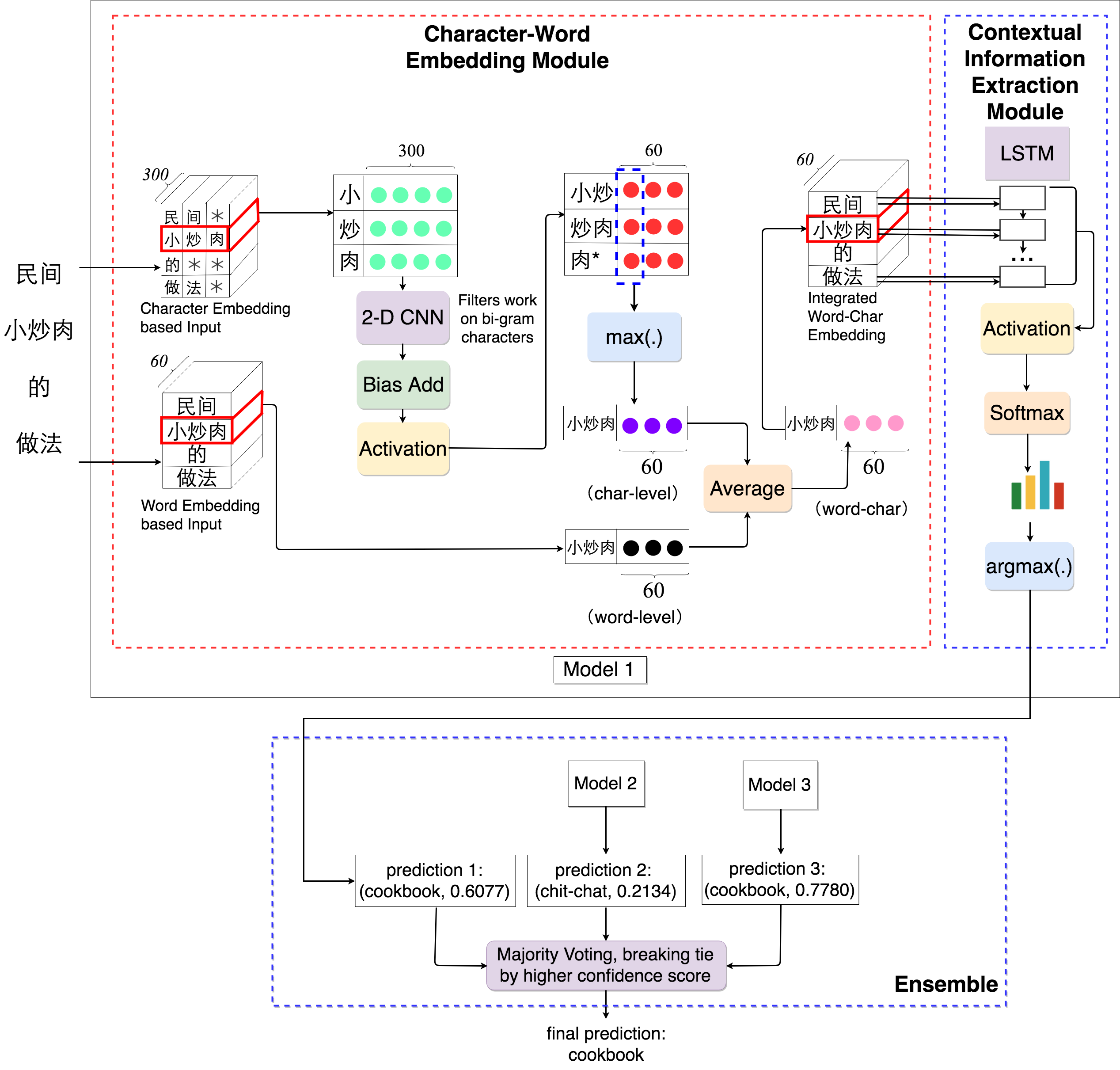}
  \caption{The proposed model architecture}
  \label{The proposed model architecture}
\end{figure}
As shown in Figure\ref{The proposed model architecture}, the overall architecture is a hybrid neural network with model ensembling at the output level. The rest of this section is on details of the component CNN based Word-Character module, the LSTM based contextual information extraction module, and the ensemble method used. 

\subsection{Word-Character Embedding Module}
This module aims to learn context-adaptive integrated Chinese word-character embeddings during training and at runtime, which does not rely on external corpus for training. \\

\noindent \textbf{Input Layer} The input layer creates placeholders for sentence-level input in both word and character representations. For a sentence of \(M\) words where each word consists of \(N\) characters (padding or truncation applied), a word-based input \(\textbf{w} \in \mathbb{R}^{M \times d_w}\) is represented as a sequence of \(M\) words, where the value of a word will be filled in by its \(d_w\)-dimensional word embedding vector. A character-based input \(\textbf{c} \in \mathbb{R}^{M \times N \times d_c}\) is a sequence of character sequences. It is a depicted as a sequence of \(M\) words, where each word is decomposed into a sequence of \(N\) characters, and the value of a character will be given by its \(d_c\)-dimensional character embedding vector. For the simplicity of notations and from the mini-batch training perspective, for a batch of \(S\) sentences, the word- and character-based inputs will be in the form of 3-D and 4-D tensors, i.e., \(\textbf{W} \in \mathbb{R}^{S \times M \times d_w}\), and \(\textbf{C} \in \mathbb{R}^{S \times M \times N \times d_c}\).\\

\noindent \textbf{Word Embedding and Character Embedding Layer}  The embedding layer takes outputs from the input layer, performs word and character embeddings look-ups, and fills the placeholders in with the corresponding word and character vectors. \\

\noindent \textbf{2-D Convolutional Layer} The purpose of this layer is to uncover the information embedded in the characters. 2-D convolutional neural networks are used to extract the features because they are good at extracting temporal and spatial information within a sequence. This layer takes the character output \(\textbf{W}\) from the embedding layer, applies a 4-D filter \(\textbf{F} \in \mathbb{R}^{1 \times V \times d_c \times d_w}\) to compute 2-D convolution operation on a window of \(V\) characters, \(d_c\) features in the second dimension, \(d_w\) features in the third dimension. In this work, we set \(V\) fixed to 2 to create bi-gram character features. For instance, an output \(o_{s,i,j,k}\) is obtained by the following equation, where \(s, i, j, k\) are in the ranges of [1, \(S\)], [\(1\), \(M\)], [\(1\), \(N\)], and [\(1\), \(d_w\)] respectively, \(\textbf{b} \in \mathbb{R}^{d_w}\) is a bias, \(f\) is a non-linear activation function applied to the convolution result plus bias:
\begin{equation}
o_{s,i,j,k} = f(\textbf{b}+\sum \nolimits_{s,i+d_i,j+d_j,q}\textbf{C}_{s,i+d_i,j+d_j,q} \cdot \textbf{F}_{d_i,d_j,q,k}) 
\end{equation}
The resulting feature map is a 4-D tensor in \(\mathbb{R}^{S \times M \times N \times d_w}\) as follows:
\begin{equation}
\textbf{O}=[o_{1,1,1,1}, \dots, o_{S,M,N,d_w}]
\end{equation}
\noindent \textbf{Max-Pooling Layer} The feature map can be interpreted as a batch of sentences in its character-level feature representations, where each word is represented by \(N\) \(d_w\)-dimensional character features. To reduce the \(N\) features to form a single most informative feature within each word, a max-pooling operator with a sliding window of \([1,1,N,1]\) is applied on \(\textbf{O}\). For example, a pooling output \(p_{s,i,1,k}\) is computed by:
\begin{equation}
p_{s,i,1,k} = \max_{1 \le j\le N} o_{s,i,j,k}
\end{equation}
Therefore, the feature map is downsampled to size \(\mathbb{R}^{S \times M \times 1 \times d_w}\).
\begin{equation}
\textbf{P} = [p_{1,1,1,1}, \dots, p_{S,M,1,d_w}]
\end{equation}
After proper reshaping, the shape of \(\textbf{P}\) should be \(\mathbb{R}^{S \times M \times d_w}\), which is in the same dimensions of word-based input \(\textbf{W}\).\\

\noindent \textbf{Integration Layer} This layer enforces the integration of the pooled character feature map with the input word vectors to bring the most out of both word and characters. Taking into consideration computation time, we average the two representations elementwisely instead of concatenating them. The result will be integrated Word-Character vectors in the dimensions of \(\mathbb{R}^{S \times M \times d_w}\).
\begin{equation}
\textbf{I} = (\textbf{W}+\textbf{P})/2.0
\end{equation}

\subsection{Contextual Information Extraction Module}
Up to this point, features at the word and character levels are extracted, but these features can not be utilized best without considering the dependency of words to their contexts. Contextual information like past neighboring words are still important to reveal the actual meaning of a word in a sentence. Thus we use an LSTM cell to remember past temporal information, and feed the integrated embedding vectors \(\textbf{I}\) as basic input to LSTM for obtaining context level features. 

The recurrent LSTM layer has 512 hidden units, and the output is stacked with a linear layer that computes an output probability distribution over the intent classes. The argmax of the distribution is taken and returned as a single model prediction.

The LSTM works as follows. An LSTM cell is accomplished by modifying the basic RNN cell, which computes the output $h$ at each timestep using both the current timestep input $\textbf{I}_t$ and the previous output $\textbf{h}_{t-1}$via 
\begin{equation}
\textbf{h}_t = \sigma(\textbf{W}_h \cdot [\textbf{h}_{t-1}, \textbf{I}_t] + \textbf{b}_h)
\end{equation}
The LSTM cell augments the RNN cell by implementing a forget gate and an input gate that control what past information is kept or discarded.
\begin{equation}
\textbf{f}_t = \sigma(\textbf{W}_f \cdot [\textbf{h}_{t-1}, \textbf{I}_t] + \textbf{b}_f)
\end{equation}
\begin{equation}
\textbf{i}_t = \sigma(\textbf{W}_i \cdot [\textbf{h}_{t-1}, \textbf{I}_t] + \textbf{b}_i)
\end{equation}
This allows the cell to have a state vector 
\begin{equation}
\textbf{C}_t = \textbf{f}_t \circ \textbf{C}_{t-1} + \textbf{i}_t  \circ \tanh(\textbf{W}_C \cdot [\textbf{h}_{t-1}, \textbf{I}_t] + \textbf{b}_C)
\end{equation}
that represents the memory of the network. The output $\textbf{h}_t$ is then calculated from the cell state $\textbf{C}$ and an output gate $\textbf{o}_t$, where
\begin{equation}
\textbf{o}_t =  \sigma(\textbf{W}_o \cdot [\textbf{h}_{t-1}, \textbf{I}_t] + \textbf{b}_o)
\end{equation}
and 
\begin{equation}
\textbf{h}_t = \textbf{o}_t \circ \tanh(\textbf{C}_t)
\end{equation}

Furthermore, to enable context-adaptive embeddings, the word and character embeddings will be updated during backpropagation.

\subsection{Model Ensemble}
The previous discussions focus on producing a single model result, to account for the variances in single model predictions, we employ a model ensemble scheme at the final output level. Following the same single model architecture as described above, we re-train three single models to obtain an ensemble, where the final prediction is selected by majority voting on single model predictions, as is also shown in Figure \ref{The proposed model architecture}.

\section{Experimental Setup}
\subsection{The SMP Dataset}
The SMP2017ECDT (SMP) dataset consists of Chinese user queries recorded from human-computer dialogues and transcribed to text \cite{Zhang2017}. It covers 31 intents including Chit-chat, Weather, Flights, Cookbook and etc. A sample query is in the format of \begin{CJK*}{UTF8}{gbsn}你好请问一下明天广州的天气如何\end{CJK*} (Hello I want to know the weather in Guangzhou tomorrow), which is labeled as the Weather intent. The dataset is split into a train set of 3,069 samples, and a test set of 667 samples. The SMP dataset is a little imbalanced as the Chit-chat category contains around 20\% of all data, whereas the rest 30 categories are distributed more evenly.

\subsection{Word and Character Embeddings}
We hypothesize that dynamically integrating character features into input word features can enhance performance, compared to using word features alone, so we use random initialization for character embeddings in our experiments, and use both randomly initialized and open-domain pre-trained word embeddings\footnote{Pre-trained word embeddings are trained on a 1G Chinese Wikipedia corpus, http://pan.baidu.com/s/1boPm2x5} for experiment and control. The character embedding vectors are initialized to be 300-dimensional with component value ranging from 0 to 1, and the word embedding vectors are initialized in the same range with a dimension of 60 to be consistent with the pre-trained word embeddings. For both randomly initialized and pre-trained embeddings, we update them during every backpropagation in training.

\subsection{Baseline Models}
Our hypothesis is that the proposed word-character based model improves intent classification accuracy compared to word-alone model, in this case the experiment is done on the hybrid Word-Character embedding based neural model and the control group is done on word embedding based LSTM, where details are presented in the previous section.

Combining the hypotheses of models and embeddings, we come to develop four sets of experiment settings, including two experiments on word-alone LSTMs, one using randomly initialized word embeddings and the other with pre-trained word embeddings, and experiments on the proposed model are also divided into two parts, one utilizes random initialization for both embeddings, the other uses pre-trained word embeddings and randomly initialized character embeddings to test out if even for pre-trained word embeddings, the proposed scheme of integrating character features can still help boost up performances. We omit experiments on using pre-trained character embeddings, because we want to lay our focus on the effectiveness of our low-effort way of generating and integrating character features dynamically, without relying on large external corpus and the need of pre-training embeddings.

For ensembles, we compare the ensemble of the proposed models to ensemble of baseline LSTMs. A comparison on our best model and the state-of-the-arts will also be drawn in the Results and Discussion section.

\subsection{Preprocessing}
To start with, Since the data is not tokenized into words, the first step is to tokenize the sentences. The Jieba Chinese tokenizer\footnote{https://github.com/fxsjy/jieba} is applied in this work. Sentences and words in sentences are then padded to fixed maximum lengths in order to do mini-batch training. Similarly, for run-time prediction, either padding or truncation to the same fixed lengths are done as a step of preprocessing. 

\subsection{Hyper-parameter Tuning}
For model selection,  we perform hyper-parameter tunings by grid search. The component single models in the ensemble share the same set of hyper-parameters.

\subsection{Evaluation Metrics}
For this multiclass intent classification problem, we measure model performance by unweighted \(F_1\) scores, implemented with the Python scikit-learn package \cite{scikit-learn}.

\section{Results and Discussion}
\begin{table*}[t!]
\centering
\begin{tabular}{ l | c}
\hline
 \textbf{Embeddings and Model} & F1(\%)\\ 
 \hline
 1 Pre-trained word, LSTM & 78.71\\
 2 Randomly initialized word, LSTM & 86.06\\
 3 Pre-trained word, randomly initialized char, Word-Char LSTM & 87.86\\
 4 Randomly initialized word and char, Word-Char LSTM & \textbf{89.51 }\\
 \hline
 \end{tabular}
\caption{A Comparison of F1 scores on the proposed models and baselines on the SMP dataset}
\label{A comparison of F1 scores on the proposed models and base lines on the SMP dataset}
\end{table*}

\begin{table*}[t!]
\centering
\begin{tabular}{ l | c}
\hline
 \textbf{Ensemble} & F1(\%)\\ 
 \hline
 Ensemble of Proposed model & \textbf{93.55}\\
 Ensemble of baseline LSTM & 87.26\\
 \hline
 \end{tabular}
\caption{Results of ensemble of the proposed model and ensemble of baseline model in unweighted $F_1$ scores (\%)}
\label{Ensemble}
\end{table*}

\begin{table*}[t!]
\centering
\begin{tabular}{ l | c | c }
\hline
 \textbf{Model} & F1 (\%) on SMP & Development time on SMP\\ 
 \hline
 The proposed single model  & 89.51  & Low \\
 Ensemble of baseline LSTMs & 87.26 & Low to medium\\
 Ensemble of the proposed models & \textbf{93.55} & Low to medium\\
 \hline
 N-gram SVM \cite{Li2017} & 90.89 & Medium, with feature engineering\\
 Ensemble of SIR-CNNs \cite{Lu2017} & 92.88 & Medium to high\\
 \begin{tabular}{@{}c@{}} Ensemble of LSTMs, domain knowledge\\ \cite{Tang2017} \end{tabular} & \textbf{93.91} & High, with feature engineering\\
 \hline
\end{tabular}
\caption{Results of our best character-level models against other methods measured by unweighted $F_1$ scores (in percentages)}
\label{Computing time}
\end{table*}

The results of the proposed Word-Char (CNN based) LSTM models and baseline LSTMs are shown in Table \ref{A comparison of F1 scores on the proposed models and base lines on the SMP dataset}. Ensemble results are given in Table \ref{Ensemble}. A Comparison on overall performance and computation time across different methods is presented in Table \ref{Computing time}.

\subsection{Effectiveness of the Word-Character Approach}
As is shown in Table \ref{A comparison of F1 scores on the proposed models and base lines on the SMP dataset}, there is an increase comparing experiment 3 to 1 or experiment 4 to 2, with an 9.15\% improvement from 1 to 3 and 3.45\% from 2 to 4. We thus verify that the CNN method as described in the Word-Character module is useful for extracting and integrating informative Chinese character level features.

Besides, comparing experiment 1 with 2, or 3 with 4, we observe an interesting finding that even though pre-trained word embeddings are dynamically updated and fine-tuned during training, the result is still worse than using randomly initialized embeddings. This is explained by the fact that the external source that pre-trains the word embeddings does not align well with the task at hand. To be more specific, in our case, the Sogou News Corpus used for pre-training does not necessarily contain contexts similar to the queries in human-computer dialogs, so for the same word, the Sogou embedding vector and the actual vector associated with the SMP task can result in quite different directions if we project them onto an embedding representation space. Thus for task-specific data, relying on pre-trained vectors can have a diminishing effect on the performance. This is especially true with Chinese data where words can have very different meanings depending on the contexts. Luckily, the Chinese characters have fewer variations and provide a more stable source to form character-level word representations, which can then be easily learned and safely used without exploiting external resources.

\subsection{Combining Word-Character Models with Ensemble Techniques}
With ensembling, the classification accuracy of ensemble of the proposed character-level models reaches 93.55\%,which gives an increase of 6.29\% compared to that of ensemble of baseline LSTMs. The ensemble method helps reduce variance and brings the best out of the constituent word-character models.

\subsection{Overall Performance Comparisons}
In this section we compare our work with the state-of-the-art works in terms of \(F_1\) scores and development time. Our ensemble model outperforms models in two of the three works and is on par with the top score model. Table \ref{Computing time} lists the scores and development times of the proposed model, the ensemble of proposed models, and the state-of-the-art works on SMP. 

The work of \cite{Li2017} uses a classical machine learning approach to text classification. They have adopted a one-vs-the-rest SVM classifier in the Lib-SVM package with n-gram character based feature vectors as input, which achieves 90.89\% \(F_1\) score. They have experimented different combinations of n-grams. In the final model, 1+2+3+4-gram vectors with a dimension of 2642 are used. Feature weights are calculated by tf-idf. The overall development time is medium. Compared to their model, our model has obtained a higher classification accuracy with a neural architecture, and is straightforward to build without feature engineering.

The work of \cite{Lu2017} has utilized pre-trained character embeddings as input and an same-structure ensemble of Self-Inhibiting Residual CNNs (SIR-CNNs). The convolution and max pooling are done in 1-D, and character embeddings are trained during training. The result gives a 92.88\% \(F_1\) score and the development time is medium to high. Our performance is better with lower computation time.

The top score comes from an ensemble model of 10 single LSTM (with multiple hidden layers) models along with data-based keyword extraction proposed by \cite{Tang2017}. They have developed a domain keyword based LSTM classifier and applied ensembling techniques to integrate 10 retrained such classifiers of the same parameters, and finally used majority voting to select the final prediction. They have trained word vectors on 10G Weibo (a Chinese microblogging website) data with fine tuning. Due to fairly complicated feature engineering, such as domain keyword extraction, the development time of their model is very high. Instead of ensembling a great number of fine-tuned complex single models with a feature extraction algorithm, the ensemble with our proposed word-character models does not require feature engineering and comprises of less and simpler constituent models, which makes faster training possible while achieving a comparable to the state-of-the-art result. Besides, their best single classifier performance on the test set is not given, so we are not able to compare their single model result and time to ours.

\section{Conclusion}
In this paper, we address the Chinese intent classification problem and propose a low-effort integrated word-character approach that enhances classification accuracy compared to models using only Chinese word embeddings. We run experiments on the SMP dataset, with different word and character embedding configurations. Our single model achieves 89.51\% on SMP. Our main findings are that the Chinese SMP data benefits more from the character approach, and we do not need to rely on pre-trained word embeddings using the proposed method. The proposed word-character module exploits the internal word and character relationships via CNN and pooling, and the embeddings are learned during training by optimizing the same loss function on logits as the word embedding model does. 

Taking into account the ensemble method, we observe an improvement from ensembles without characters to those with characters, and the best ensemble achieves 93.55\% on SMP, which are on par with the state-of-the-art. The proposed model is easy to implement and train, which greatly reduces the development time compared to works that rely on feature engineering and sophisticated architectures.

Our future work could focus on conducting experiments on different pooling strategies and embedding combining methods, for example, instead of giving equal weights to the word and character embeddings when combining, we would like to find out the contributions of  word and character embeddings by experimenting on various weights. In addition, concatenation instead of weighted average could be used. Another focus of our work will be exploring different ensembling and stacking techniques with the character-level models. Last but not least, we will investigate attention mechanisms that could potentially further improve intent classification results.

\bibliography{cloudminds}
\bibliographystyle{acl}

\end{document}